\documentclass{article}
\usepackage{ijcai25}

\usepackage{times}
\usepackage{soul}
\usepackage{url}
\usepackage[hidelinks]{hyperref}
\usepackage[utf8]{inputenc}
\usepackage[small]{caption}
\usepackage{graphicx}
\usepackage{amsmath}
\usepackage{amsthm}
\usepackage{booktabs}
\usepackage{algorithm}
\usepackage{algorithmic}
\usepackage[switch]{lineno}
\usepackage{pdfpages}
\usepackage{subcaption}
\usepackage{enumitem}
\newcommand{\anon}[1]{[anonymous]}

\title{Who Benefits from AI Explanations? Towards Accessible and Interpretable Systems}

\author{
Maria J. P. Peixoto$^{1,2}$
\and
Akriti Pandey\and
Ahsan Zaman\And
Peter R. Lewis$^{1,3}$\\
\affiliations
$^1$Ontario Tech University\\
\emails
\{$^2$mariajoelma.pereirapeixoto, $^3$peter.lewis\}@ontariotechu.ca,
p.akriti@yahoo.com,
ahsan.zamn1@gmail.com
}

\begin{document}

\maketitle

\begin{abstract}
    As AI systems are increasingly deployed to support decision-making in critical domains, explainability has become a means to enhance the understandability of these outputs and enable users to make more informed and conscious choices. However, despite growing interest in the usability of eXplainable AI (XAI), the accessibility of these methods, particularly for users with vision impairments, remains underexplored. This paper investigates accessibility gaps in XAI through a two-pronged approach. First, a literature review of 79 studies reveals that evaluations of XAI techniques rarely include disabled users, with most explanations relying on inherently visual formats. Second, we present a four-part methodological proof of concept that operationalizes inclusive XAI design: (1) categorization of AI systems, (2) persona definition and contextualization, (3) prototype design and implementation, and (4) expert and user assessment of XAI techniques for accessibility. Preliminary findings suggest that simplified explanations are more comprehensible for non-visual users than detailed ones, and that multimodal presentation is required for more equitable interpretability.
\end{abstract}

\section{Introduction}


A phenomenal worldwide effort aims to unlock the benefits of AI across society, business, and the economy. As a result, important questions arise around the responsible use and trustworthiness of AI, and further, about the equity and power implications that stem from the widespread use of AI technologies. Therefore, one of the key requirements of an AI tool today is explainability: as citizens, we need to have the right and ability to know when a machine makes a prediction or decision concerning us that may be biased, or simply based on a characteristic or factor we believe to be unfair or would be inappropriate or illegal. This includes the ability to interrogate the system as to what it is doing; why it is doing that; how it does it; and why it does it this way \cite{9445758}. Unfortunately, even when AI systems provide explanations for their decisions, these are often not accessible to persons with sight loss.

The term `accessible' in this paper refers to removing or minimizing barriers that may affect persons with disabilities in accessing, comprehending, participating in, and effectively using opportunities, spaces, and technologies such as AI systems \cite{GOLDENTHAL2021106975,KULKARNI201991}. AI systems can offer users greater autonomy by increasing transparency and explainability. By better understanding the recommendations or decisions generated by AI, users can make more informed and conscious choices, using their critical judgment to accept or reject the suggestions \cite{MILLER20191}. The awareness of the AI decision-making process is especially important in areas where AI decisions have significant real-life implications, such as healthcare, vehicle automation, or the judicial system. With a clearer understanding of the processes behind AI decisions, users can avoid unwarranted trust on the technology. Instead, they can develop a more appropriate level of trust corresponding to the AI's performance and the context in which it operates, assessing its strengths, limitations, and potential biases or risks.


In recent years, various laws, proposals, and action plans have been introduced to enhance transparency, fairness, accessibility, and equity in automated systems, especially those based on AI. Examples include the European Union's Digital Services Act (DSA) \cite{europaEUsDigital} and AI Act \cite{Eu_AIAct}, the National Institute of Standards and Technology (NIST) AI Risk Management Framework \cite{nistRiskManagement} in the United States, Canada's Accessible and Equitable Artificial Intelligence Systems \cite{canadaAccessibleEquitable} and the Artificial Intelligence and Data Act (AIDA) \cite{CanadaAIDA2024}, and, in Brazil, the approved bill that establishes a legal framework for artificial intelligence \cite{camaraBrasil}. Despite these initiatives, various counterefforts hamper progress. For instance, companies claim that revealing their systems' operations could violate trade secrets, enable reverse engineering, or facilitate fraud \cite{GROZDANOVSKI2025106117}. Moreover, some governments have resisted measures promoting responsible AI, as evidenced by the repeal of Executive Order 14110 of October 30, 2023 (Safe, Secure, and Trustworthy Development and Use of Artificial Intelligence) in the United States \cite{whitehouseInitialRescissions}. Meanwhile, the need for regulations and legislation mandating transparency and explainability in AI systems has become increasingly urgent. In the absence of such policies, organizations can operate without disclosing the internal workings of their algorithms, potentially leading to negligence in developing systems that do not prioritize ethical and fair practices.

Additionally, although there is increasing attention and interest in making AI more transparent and accessible, as discussed above, very few studies in the recent literature address the accessibility of explainable artificial intelligence, as evidenced by the systematic literature survey in \cite{nwokoye2024}. In light of these regulatory and ethical challenges, this paper responds to the pressing need for more inclusive and comprehensible AI systems by focusing on the accessibility of AI explanations. Specifically, our contributions include a literature review that examines which eXplainable AI (XAI) techniques are evaluated with end users and whether these evaluations include users with sight loss. This review identifies a gap concerning the inclusion of persons with disabilities and the development of accessible explanations (e.g., non-visual alternatives to graphs, diagrams, and tables). Additionally, we present a methodological proof of concept that demonstrates, on a controlled scale, the feasibility of integrating XAI techniques into real-world scenarios in a manner accessible to persons with visual impairments. This proof of concept aims to inform and encourage further research and practice, as well as to provide recommendations for the XAI community on creating more inclusive explanations.

Our methodological proof of concept consists of four components: (1) a categorization of AI systems to guide the development of case study scenarios, (2) a framework for defining personas that captures key attributes within the created case study context, (3) a functional prototype that implements XAI techniques, considering both the case study and persona previously defined, and (4) an evaluation process involving expert consultations and a co-design session with users who have lived experience of sight loss. To support our contributions, this paper is grounded in two main research questions:

\begin{enumerate}
    \item Are there any gaps or challenges related to accessibility in current XAI approaches?
    \label{Q1}
    \item How can we evaluate the comprehension of AI explanations among non-visual users?
    \label{Q2}
\end{enumerate}

The remainder of this paper is structured as follows. In Section \ref{background}, we review the current state of accessibility challenges in eXplainable AI (XAI) and present the results of our literature review, emphasizing the research gaps in accessible XAI methods. Section \ref{methodology} outlines our four-component methodological proof of concept designed to explore how AI explanations are perceived by non-visual users. Finally, Section \ref{conclusion} summarizes our findings and suggests directions for future research in accessible and explainable AI.

\section{Literature Review in XAI}
\label{background}


According to the World Health Organization (WHO) in 2023 \cite{WHO2024}, an estimated 16\% of the global population experiences significant disabilities, including dementia, blindness, and spinal cord injuries. Another WHO report on ``Blindness and Vision Impairment" \cite{WHO_Blindness2024} estimates that at least 2.2 billion people worldwide have some form of near or distance vision impairment. Given these statistics and the increasing presence of AI-driven systems in daily life, efforts should be made to reduce barriers and ensure that individuals with sight loss and other disabilities can access and participate in the design and development of AI-based systems, particularly concerning their ability to understand the decisions and recommendations provided by these applications. However, the sight loss community has been largely overlooked when it comes to access to AI explanations \cite{nwokoye2024}. The article \cite{nwokoye2024} claims that research on the accessibility of XAI techniques is nearly nonexistent, and many XAI methods rely on visual explanations.

Relying on visual explanations creates a significant barrier to inclusivity and accessibility in AI interpretability. The lack of inclusive XAI tools not only prevents persons with disabilities from examining and contesting biased AI decisions but also limits their participation in discussions about AI ethics and governance. In response to these concerns and to address question \ref{Q1} of this research, we conducted a literature review to investigate whether XAI techniques are evaluated with end users, which techniques are most commonly adopted, what forms of accessibility they include, who these users are (their backgrounds, accessibility needs, education levels, AI expertise, and AI literacy), and whether there is any assessment of how well users understand AI explanations.

\begin{figure}
    \centering
    \includegraphics[width=1\linewidth]{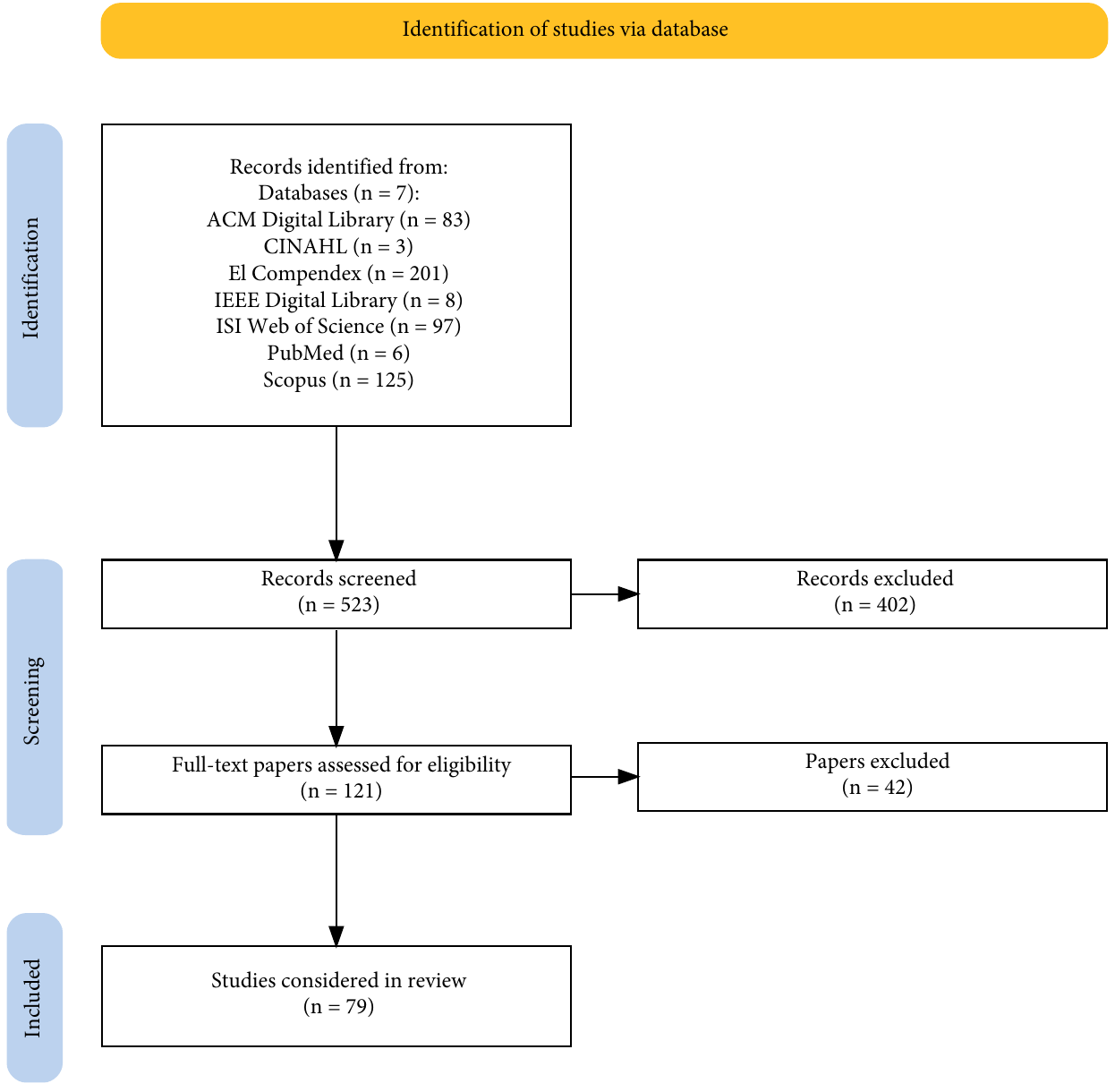}
    \caption{PRISMA Flow Diagram}
    \label{fig:prisma}
\end{figure}

To investigate recent literature, we conducted a review of publications from 2019 to 2024 using the search string (XAI OR ``AI explanation") AND (end-user OR user) AND (evaluation OR validation). Figure \ref{fig:prisma} presents the PRISMA \cite{haddaway2022prisma2020} flow diagram illustrating our methodology. After removing duplicates, we identified 523 articles. We then screened their titles and abstracts to narrow the selection to 121 papers, which were subsequently analyzed individually, resulting in a final set of 79 papers for our study.

We applied the following exclusion criteria: papers that were literature reviews or meta-reviews, studies that evaluated XAI techniques exclusively with developers, those in which XAI techniques were not evaluated with end users, and papers that did not focus on evaluating XAI techniques.

Table S1 in the Supplementary Material encompasses a broad range of explainable AI (XAI) techniques evaluated across multiple studies. Traditional, well-established methods, such as SHAP and LIME, appear frequently, being cited in 33 and 30 out of the 79 analyzed papers, respectively. Other approaches, such as Grad-CAM or CAM, along with counterfactual explanations, are discussed in 9 out of the 79 papers, while decision trees appear in 6. Additionally, the table includes 23 novel methods, indicated by a ``+" symbol. This variety reflects the active research landscape in XAI, where methods are tailored to different data modalities (such as images, tabular data, and text) and application areas (for example, healthcare, and fraud detection). The studies employ a variety of evaluation methods, many of which are user-focused, involving surveys \cite{arai_enhancing_2024,longo_study_2024,metta_advancing_2024}, questionnaires \cite{van_der_waa_evaluating_2021,kim_designing_2023,del_aguila_escobar_oboe_2024}, interviews \cite{aliyeva_uncertainty-aware_2024,martalnez_designing_2023,wolfel_user_2022}, task-based experiments \cite{nowaczyk_evaluation_2024,schmude_impact_2023}, and even eye-tracking \cite{meza_martinez_does_2023} research to assess user interaction and understanding. Furthermore, some studies combine algorithmic validations with human-centered evaluations \cite{guo_mimicri_2024,aliyeva_uncertainty-aware_2024}. For example, they may use multi-criteria decision analysis with Z-numbers, conduct simulated experiments, and gather user assessments.

Evaluations involve different user groups. Some studies (29 out of 79) focus on domain experts, such as healthcare professionals, neurosurgeons, radiologists, air traffic controllers, and fraud analysts, while others (33 out of 79) involve general or non-expert users, including laypersons, high school students, and crowd-sourced participants. In certain cases, both technical and non-technical users are included (17 out of 79 papers analyzed). This range of participants highlights that the effectiveness of an XAI technique often depends on the user’s background and domain knowledge. Studies marked with a ``+" indicate those that propose new XAI techniques, such as ACE, I-CEE, PKiL, and LIME+. Understanding how different user groups interact with these techniques is important for tailoring explanations that meet their specific needs and enhance overall comprehension.

SHAP (33 out of 79 papers analyzed) and LIME (30 out of 79) are the most frequently evaluated techniques in Table S1 in the Supplementary Material. Their widespread use likely stems from their model-agnostic nature, which enables them to be applied across different models, and from their ability to offer local and global explanations by assigning importance to input features. While SHAP provides theoretically robust attributions, it often visualizes them as complex bar plots or summary plots, which may not translate well to auditory or tactile formats. LIME, though computationally lighter, similarly depends on visual elements such as weighted feature lists or graphs. Other commonly used methods include Grad-CAM or CAM, counterfactual explanations, and decision trees. Grad-CAM and CAM relies heavily on visual heatmaps for interpretability in image-based tasks, and counterfactual explanations often use comparative visuals to illustrate minimal changes needed to alter outcomes. Although more structurally transparent, decision trees and rule-based systems also tend to present their outputs graphically. This overall emphasis on visual representation creates significant barriers for users with vision impairments. These tendencies highlight an accessibility gap in current XAI practices, where interpretability is often synonymous with visualization, excluding users relying on non-visual modalities. Therefore, the choice of explanation technique must consider not only the model and domain but also the accessibility needs of diverse users.

Assessing user understanding of XAI techniques is crucial because the goal of XAI is not only to provide correct attributions but also to enhance users’ mental models of AI systems. If users do not understand the explanations, effective decision-making may not be achieved. To measure user understanding, many studies have used surveys and questionnaires with Likert scales to gauge perceived understandability, trust, and satisfaction. In task-based evaluations, users are asked to make decisions based on the provided explanations, and metrics such as decision accuracy, decision time, and error rates are recorded. Some experiments use pre-test and post-test designs to assess learning gains after exposure to the explanations. Additionally, behavioral measures like eye-tracking help researchers understand how users interact with the explanations, while other evaluations include assessments of cognitive load and changes in the mental model using direct tests like Cloze tests.

The Table S1 in the Supplementary Material includes studies with a wide range of participants, from domain experts like radiologists and neurosurgeons to non-experts and general users. This diversity influences evaluation results. Expert users tend to demand more detailed and technically accurate explanations; they scrutinize the fidelity of the explanations and notice nuances that non-experts might overlook. In contrast, non-expert users may prefer simplicity, clarity, and minimal cognitive load, and overly technical explanations may lead to confusion or mistrust. Therefore, an XAI technique that works well for experts might not translate effectively to a lay audience, and vice versa \cite{arai_enhancing_2024,morrison_impact_2024,jansen_contextualizing_2024}. The intended target user should guide the choice of XAI method. For example, clinical decision support systems require methods that resonate with medical professionals, while consumer-facing applications may benefit from more intuitive, simplified explanations. Evaluations that include both expert and non-expert users can provide insights into adapting explanations for mixed audiences, potentially broadening adoption.

Some studies in Table S1 in the Supplementary Material, identified with an asterisk ``*", mention accessibility concerns but focus primarily on color blindness, approached through colorblind-friendly palettes and design principles. Among the works analyzed, three papers \cite{longo_study_2024,nagy_interpretable_2024,veldhuis_explainable_2022} out of 79 discuss these concerns, and only \cite{longo_study_2024} includes participants reporting some level of visual impairment in its evaluation of XAI techniques. In \cite{longo_study_2024}, participants self-assess whether their impairment has no effect, a moderate effect, or a minor effect on their vision. However, the paper offers few details about the measures taken based on these assessments. The authors merely note that ``among the participants, no bias due to visual impairment could be identified" \cite{longo_study_2024}. Examining the ``participant expertise" column in Table S1 in the Supplementary Material reveals a variety of backgrounds and expertise, but no other studies include individuals with sight loss or other disabilities. This gap underscores the need for more inclusive research methods that actively involve people with diverse disabilities, ensuring that XAI’s benefits reach all users.

\section{Methodological Proof of Concept}
\label{methodology}
To explore how AI explanations are understood by non-visual users, as posed in research question \ref{Q2}, we designed a methodological proof of concept that operationalizes this inquiry through a structured, four-part approach. First, we developed a categorization scheme for AI systems to inform the construction of meaningful case study scenarios. Second, we created a detailed user persona to reflect a lived experience of persons with sight loss, grounding the scenario in real-world accessibility challenges. Although we recognize the diversity and complexity of disability experiences, especially the many forms and degrees of sight loss, this study adopts a single illustrative persona as an initial reference point. The intention is not to represent all possible user profiles but to provide a concrete basis for scenario development and accessibility analysis. This approach serves as a starting point, with the understanding that future research should incorporate a broader range of personas to capture the multifaceted nature of disability. Third, we designed and implemented an interactive prototype equipped with XAI techniques tailored to the selected scenario and user profile. Fourth, this methodology was evaluated through expert consultations and a co-design session involving users with lived experience of sight loss. The following subsections describe each component of the methodology in detail.

\subsection{Categorization of AI Systems}
This subsection introduces a categorization of AI systems designed to support the development of the case study scenarios further discussed. By organizing AI across dimensions such as impact, functionality, transparency, and reasoning, we provide the basis for a comprehensive framework that underpins the design of these scenarios. This categorization represents the synthesis of the types of AI applications that people may encounter in everyday life, whether for sports, study, work, or leisure. This synthesis integrates four central questions about these systems' decision-making: who is affected, for what purpose the decision is made, how transparent the system is, and what reasoning leads to the decision. We drew on the frameworks ``AI Use Taxonomy: A Human-Centered Approach" \cite{theofanos2024ai} and the ``OECD Framework for the Classification of AI Systems" \cite{oecd2022oecd} to ground these questions and define the dimensions.

This structured approach enables an understanding of the challenges posed by different types of AI and supports the selection of appropriate eXplainable AI (XAI) methods. Table S2 in the Supplementary Material outlines the primary dimensions considered: the audience affected by AI decisions, the specific capabilities and tasks the systems perform, the degree of transparency in their decision-making processes, and the forms of reasoning they utilize. The table provides definitions and representative examples for each category, serving as a foundational reference for the subsequent development and evaluation of the proposed prototype.

After creating Table S2 in the Supplementary Material, the categories from the sections on Impact-Centric AI, Function-Based AI, Transparency-Based AI, and Reasoning-Based AI were combined to establish the context for the case study scenario. The sections categories are not mutually exclusive, meaning they can overlap when defining a case study. Therefore, multiple categories from the same section may apply to a particular scenario. For example, a specific case study might include both ``generative" and ``task-based" categories under Function-Based AI. However, only the ``task-based" category may be emphasized as particularly relevant to the analysis. For this research, we developed and analyzed a case study based on a randomly selected combination of categories: Third-Party AI + Descriptive AI + Black-Box AI + Deductive Reasoning. Based on this combination, we have chosen to contextualize the case study within an urban traffic management scenario as follows: ``The traffic department of a large city utilizes an AI-based system to efficiently manage urban traffic, optimizing vehicle flow and reducing congestion. This system collects and analyzes real-time traffic data, providing a comprehensive overview of road conditions, intersections, and congestion points on a dashboard. It can also predict potential future congestion by analyzing the current traffic state. Based on these predictions, the system generates recommendations for adjusting traffic light programming, creating alternative routes to reduce congestion, and modifying speed limits. All suggestions comply with local traffic laws and regulations. After recommendations are generated, the traffic manager evaluates them to decide whether to implement the proposed changes''.

\subsection{Persona Definition and Contextualization}
Additionally, we developed a detailed persona to represent a worker within the scenario described above. This persona was created in collaboration with experts from the Canadian National Institute for the Blind (CNIB), including individuals with lived experience of sight loss. Personas capture the diverse range of users who might interact with the prototypes in this study \cite{coorevits2016bringing} and are crucial for understanding user behavior, contextualizing their interactions, and identifying potential barriers. Ultimately, these personas guide the design process to ensure that the prototype is user-centered and tailored to meet the specific needs, preferences, and challenges of user groups:

\begin{itemize}[label={}]
    \item \textbf{Name:} Caroline
    \item \textbf{Age:} 45 years
    \item \textbf{Occupation:} Traffic manager - Professional responsible for supervising and planning vehicle and pedestrian flow on urban roads and highways, minimizing congestion, and promoting efficient mobility.
    \item \textbf{Work technologies:} She uses a system (our prototype) to monitor and track traffic in her region, covering some cities in real-time. The prototype to be presented is an illustrative example. This software employs machine learning (ML) algorithms to analyze data from city sensors and predict traffic flow.
    \item \textbf{Visual Disability:} Total blindness since birth
    \item \textbf{Challenging to interact with the system:} Caroline's primary challenge is ensuring the system is accessible to non-visual users. She needs the information to be presented auditory, allowing her to correctly interpret traffic data and make informed decisions.
    \item \textbf{Method of interacting with the system:} Caroline's primary interface with the system is a computer in the traffic control center. She relies on a screen reader for text-to-speech conversion. When she encounters any difficulties using the system, Caroline asks the company's support team for help.
    \item \textbf{System limitation:} System explanations related to flow prediction are mostly figures or graphs.
    \item \textbf{User background:} Caroline is familiar with the system input data, such as detector ID, location, speed, time, and occupancy, and she expects output values for traffic flow prediction. Knowing the estimated flow, she can pinpoint potential congestion areas. Also, she is interested in how the ML model makes predictions and can use the Explainable Artificial Intelligence (XAI) techniques provided by the system to verify the system's decision motivations.
    \item \textbf{Interaction example:} 
    \begin{enumerate}
        \item Using the provided prototype, Caroline clicks on the ``Get Traffic Flow Prediction and Location Details'' button to initialize the system, which receives real-time data from several sensors spread throughout the region where Caroline monitors traffic.
        \item After that, a table is displayed with data related to predicted flow, city name, the identification of the sensor used to collect the information, the average speed on the road, and the occupancy rate.
        \item Caroline can then choose which XAI technique to use to understand how the system's algorithm works to predict the flow. The XAI options available in the system are: LIME - Simplified Explanation, LIME - Detailed Explanation, SHAP - Simplified Explanation and SHAP - Detailed Explanation.
        \item After selecting the XAI method, information appears on the screen, and Caroline analyzes it to obtain insights into the model's functioning.
    \end{enumerate}
\end{itemize}

\subsection{Prototype Design and Implementation}
\begin{figure}
    \centering
    \includegraphics[width=1\linewidth]{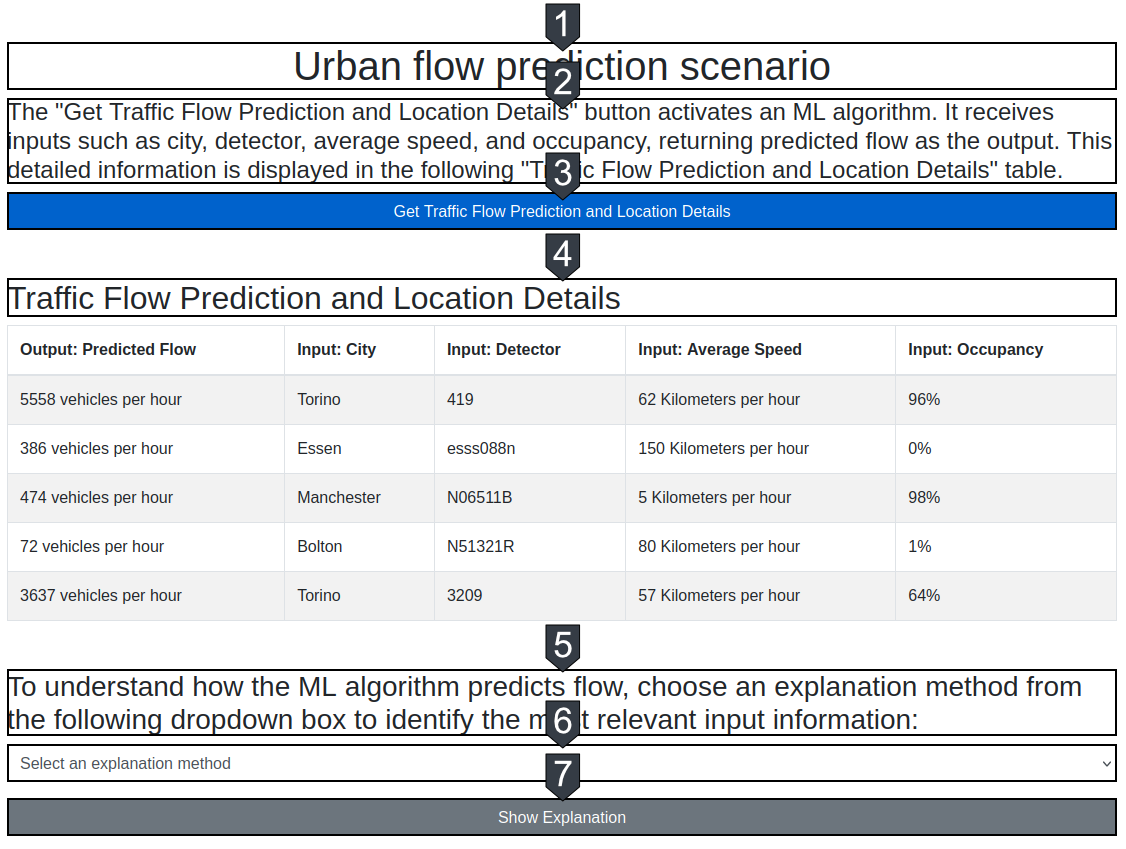}
    \caption{Prototype created for the urban traffic scenario}
    \label{fig:prototype}
\end{figure}

Based on the described case study and persona, we then developed an AI-driven web application prototype for an urban flow prediction scenario. This prototype, illustrated in Figure \ref{fig:prototype}, is designed to be compatible with screen readers to support navigation. The system navigation flow is numbered 1 to 7, as shown in Figure \ref{fig:prototype}. During the development phase, different screen readers, including JAWS\footnote{\url{https://www.freedomscientific.com/products/software/jaws/}}, NVDA\footnote{\url{https://www.nvaccess.org/download/}}, and Orca Screen Reader\footnote{\url{https://help.gnome.org/users/orca/stable/index.html.en}}, were utilized and tested on both Windows and Ubuntu operating systems.

We developed the prototype shown in Figure \ref{fig:prototype} as a web-based application, combining an interactive front-end built with HTML, Bootstrap, jQuery, and Plotly for enhanced interaction, visualization, and accessibility, along with a Flask-based back-end for ML inference and explanation generation. To predict vehicle flow, we utilized the UTD19 dataset (titled UTD19.csv) \cite{loder2019understanding}, which we partitioned into separate training and inference sets. We trained a Random Forest regression model from scikit-learn, configured with 100 estimators, on this dataset. The trained model was then saved and later loaded into our Flask application together with the inference dataset to perform the predictions. The columns we utilized from the UTD19.csv dataset were flow, interval, occupancy, speed, day, detid, and city.

The XAI techniques we selected for implementation and user evaluation in this research were LIME (Local Interpretable Model-Agnostic Explanations) \cite{lime} and SHAP (Shapley Additive Explanations) \cite{shap}. These techniques are widely recognized as state-of-the-art tools for explaining AI system decisions across various domains and user backgrounds \cite{app14198884}. The prototype features LIME - Simplified Explanation (\ref{fig:sub1}) and SHAP - Simplified Explanation (\ref{fig:sub3}), which present bar charts illustrating the importance of different features in the predictions. Additionally, it includes LIME - Detailed Explanation (\ref{fig:sub2}) and SHAP - Detailed Explanation (\ref{fig:sub4}), offering more detailed insights beyond feature importance. Algorithm \ref{alg:traffic_xai} presents a pseudocode of the prototype:

\begin{algorithm}[tb]
    \caption{Traffic-Flow Prediction}
    \label{alg:traffic_xai}
    \textbf{Input}: UTD19.csv, pretrained model\\
    \textbf{Parameter}: XAI method $m \in \{\text{LIME},\text{SHAP}\}$\\
    \textbf{Output}: Prediction table and explanations\\
    \begin{algorithmic}[1]       
        \STATE \textbf{Import} lib imports
        \STATE $M \leftarrow$ \textsc{load}(\text{``pretrained\_model''})
        \STATE $(D, X) \leftarrow$ \textbf{LoadAndProcessData}()
        \STATE $C_{\text{xai}} \leftarrow$ empty dictionary   
        \vspace{4pt}

        \STATE \textbf{function} \emph{LoadAndProcessData}()
        \STATE \quad $D \leftarrow$ read\_csv(\text{``UTD19.csv''})
        \STATE \quad $X \leftarrow D[\text{interval, occ, speed}]$ as float32
        \STATE \quad $D[\text{flow\_pred}] \leftarrow M.\textsc{predict}(X)$
        \STATE \quad \textbf{return} $(D, X)$
        \STATE \textbf{end function}
        \vspace{2pt}

        \STATE \textbf{function} \emph{PredictionTable}($D$)
        \STATE \quad $L \leftarrow$ empty list
        \FORALL{$r$ in $D$}
            \STATE \quad append \{PredFlow, City, Detector, Speed, Occupancy\} to $L$
        \ENDFOR
        \STATE \quad \textbf{return} $L$
        \STATE \textbf{end function}
        \vspace{2pt}

        \STATE \textbf{function} \emph{ProcessExplanation}($m$, $X_\ast$)
        \STATE \quad $k \leftarrow$ hash($m, X_\ast$)
        \IF{$k \in C_{\text{xai}}$}
            \STATE \quad \textbf{return} $C_{\text{xai}}[k]$
        \ENDIF
        \IF{$m$ starts with LIME}
            \STATE \quad $r \leftarrow$ LimeExplanation($M, X_\ast$, variant($m$))
        \ELSE  
            \STATE \quad $r \leftarrow$ ShapExplanation($M, X_\ast$, variant($m$))
        \ENDIF
        \STATE \quad $C_{\text{xai}}[k] \leftarrow r$
        \STATE \quad \textbf{return} $r$
        \STATE \textbf{end function}
        \vspace{6pt}
    \end{algorithmic}
\end{algorithm}

\subsection{Expert and User Assessment of XAI Techniques for Accessibility}
\label{evaluation}
This study was performed in compliance with the regulations of Ontario Tech University’s Research Ethics Board (REB) Committee under file number 17953, approved on Oct 7, 2024. Both expert and user evaluations, as described below, were carried out through online sessions lasting one hour each. These sessions were recorded and transcribed for subsequent analysis and review.

\subsubsection{Expert Evaluation}
\begin{figure}
    \centering
    \begin{subfigure}[c]{0.48\textwidth}
        \centering
        \includegraphics[width=\textwidth]{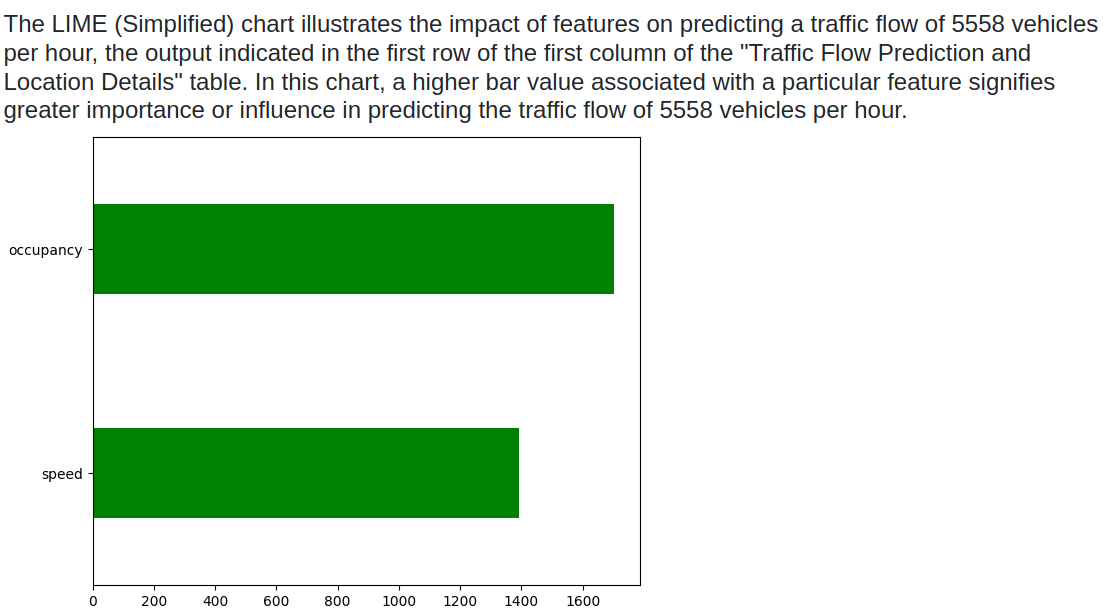}
        \caption{LIME - Simplified Explanation}
        \label{fig:sub1}
        \vspace{2em}
    \end{subfigure}
    \begin{subfigure}[c]{0.48\textwidth} 
        \centering
        \includegraphics[width=\textwidth]{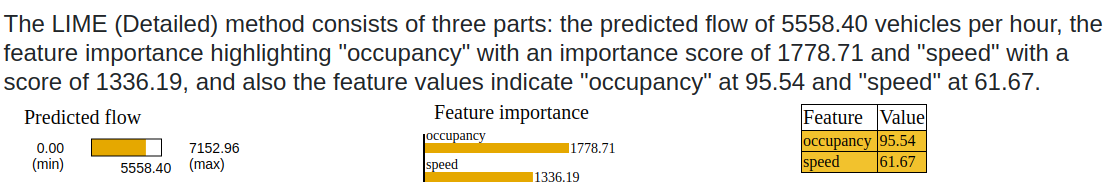}
        \caption{LIME - Detailed Explanation}
        \label{fig:sub2}
        \vspace{2em}
    \end{subfigure}
    \begin{subfigure}[c]{0.48\textwidth}
        \centering
        \includegraphics[width=\textwidth]{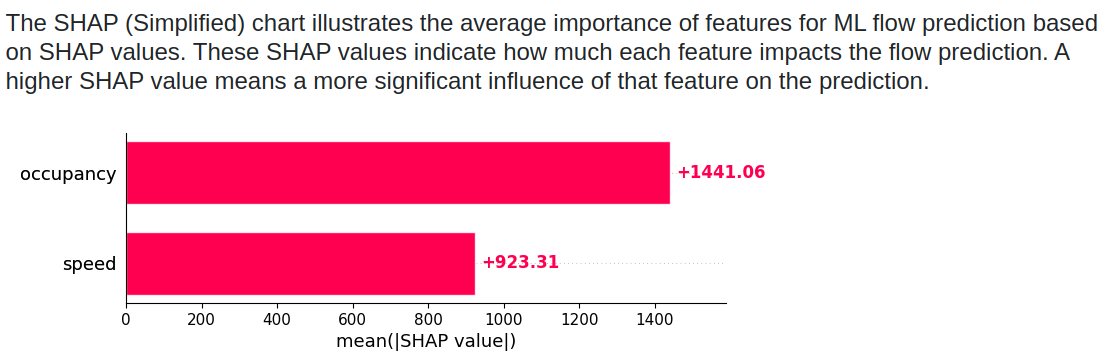}
        \caption{SHAP - Simplified Explanation}
        \label{fig:sub3}
        \vspace{2em}
    \end{subfigure}
    \begin{subfigure}[c]{0.48\textwidth}
        \centering
        \includegraphics[width=\textwidth]{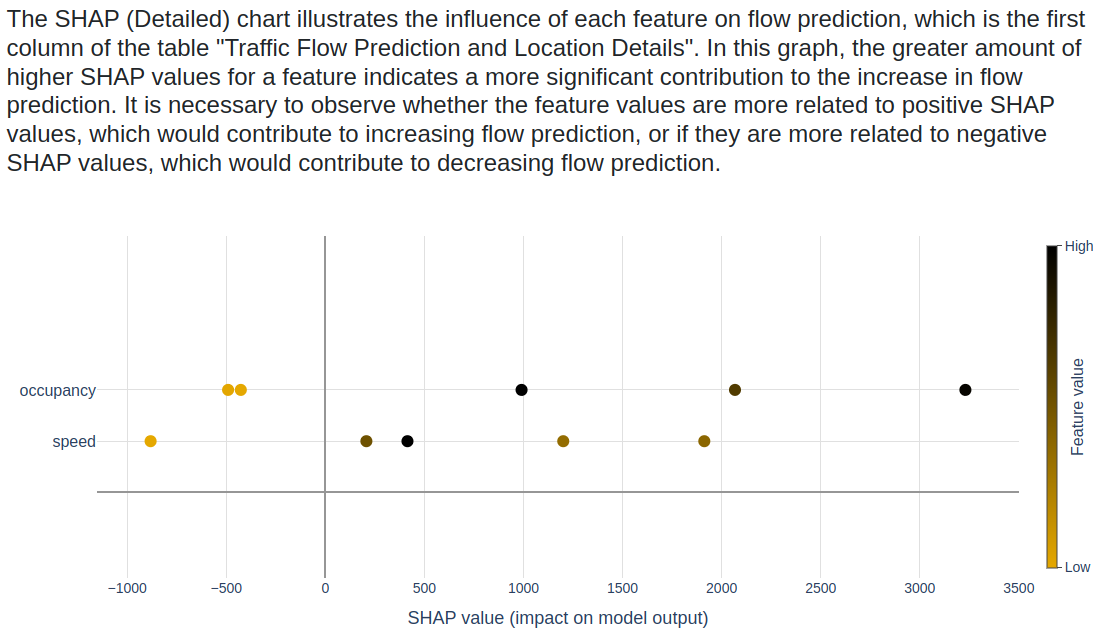}
        \caption{SHAP - Detailed Explanation}
        \label{fig:sub4}
    \end{subfigure}
    \caption{XAI techniques available in the prototype.}
    \label{fig:XAI_geral}
\end{figure}

The XAI techniques demonstrated in the prototype were initially evaluated during a research cluster meeting involving six accessibility experts. These experts, who included accessibility researchers, testers, evaluators, and consultants from the Canadian National Institute for the Blind (CNIB), as well as persons with lived experience of sight loss, possessed a general understanding of AI and XAI. They were asked to assess the accessibility of the prototype based on their own professional expertise and lived experience. They were presented the XAI methods and also asked to provide suggestions for improvement. Their primary recommendation was to include descriptive text accompanying each explanation, as illustrated in Figure \ref{fig:XAI_geral}. Additionally, during the prototype's development, the Accessible Rich Internet Applications (WAI-ARIA) \cite{world2023accessible} suite of web standards was followed to ensure accessible interactions, enabling screen readers to describe images generated using XAI techniques.

\begin{figure}
    \centering
    \includegraphics[width=1\linewidth]{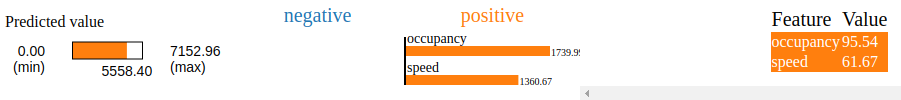}
    \caption{Default explanation generated from LIME - Detailed explanation}
    \label{fig:lime_default}
\end{figure}

The experts also evaluated the information presentation and colour contrast used. For example, the default explanation for LIME - Detailed Explanation initially included a ``negative" label on the feature importance graph, even when no features had negative contributions (Figure \ref{fig:lime_default}). We have corrected this issue, as illustrated in Figure \ref{fig:sub2}. Additionally, we enhanced the SHAP - Detailed Explanation by implementing keyboard navigation and sound-based feedback for each data point.

\begin{figure}
    \centering
    \includegraphics[width=1\linewidth]{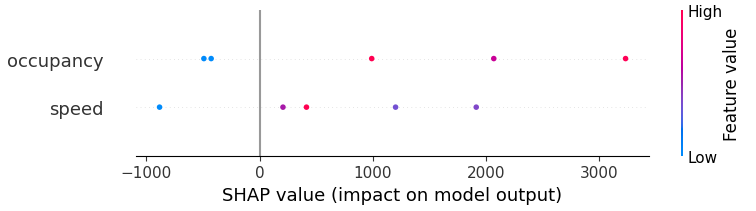}
    \caption{Default explanation generated from SHAP - Detailed Explanation}
    \label{fig:shap_default}
\end{figure}

Regarding colour contrast, we made adjustments to the LIME - Detailed Explanation and SHAP - Detailed Explanation techniques (respectively Figures \ref{fig:sub2} and \ref{fig:sub4}) to improve high contrast accessibility compared to their default versions (Figures \ref{fig:lime_default} and \ref{fig:shap_default}).

\subsubsection{User evaluation}
To evaluate the explanations provided by the AI-based prototype, ten users with lived experience of sight loss signed up to participate in an online co-design session held on November 19, 2024. However, only four participants attended. One participant experienced technical difficulties and had to leave, resulting in three active participants. Among the participants, two were completely blind, and one had low vision, utilizing a magnifier. Two accessed the session through computers, while one used a smartphone. Two participants were from Ontario, and one was from Quebec, Canada. All participants were over 19 years old and had general knowledge about AI.

The session began with participants agreeing to a consent form, which was read aloud. Following this, we introduced fundamental concepts of AI. After the introduction, we presented the prototype and explained its scenario and the persona used in the study. Participants were then encouraged to interact with the system and share their feedback regarding the explanations provided for traffic flow predictions. To guide the discussion, we posed the following questions: (1) What information and insights did you extract from each method? Consider the methods separately, starting with LIME – Simplified, then LIME – Detailed, followed by SHAP – Simplified, and finally SHAP – Detailed; (2) What challenges do you think the persona Caroline faces in accessing and comprehending the information provided by the XAI methods?; (3) What improvements could be implemented to enhance the accessibility and relevance of these XAI methods for the persona Caroline?; and (4) Among the XAI methods available in the prototype, which do you believe the persona Caroline would prefer, and why?

One participant mentioned having difficulty understanding the explanation from the SHAP - Detailed Explanation method, even with the additional descriptions provided. Another participant pointed out difficulties with the table presented in the LIME - Detailed Explanation approach. They noted that, although the information was readable, its significance was often unclear for users who are completely blind. All three participants agreed that while the detailed explanations provided more information, the simplified explanations offered a better overall understanding.

\subsection{Possible recommendations and insights}
The evaluations conducted with CNIB experts and users with sight loss emphasize the necessity for further studies to assess the effectiveness of state-of-the-art XAI techniques in providing accessible explanations. This research remains preliminary and does not establish generalized recommendations or patterns for developing accessible AI system explanations.

Nevertheless, some insights emerged, highlighting the necessity for explanations in varied formats tailored to users' backgrounds and abilities. Participants particularly appreciated the fact that LIME and SHAP provided different, complementary information. They expressed a preference for explanations presented in paragraphs or simplified point form rather than complex charts. Specifically, linear charts were found to be more accessible than tabular formats with columns and rows. Even when row headers were read aloud, users found it challenging to comprehend and effectively picture tabular data.

Although detailed versions of explanations provided more information, participants noted that an overview was preferable for quick comprehension, whereas detailed LIME explanations were favored when in-depth understanding was required. A single model of explanation, typically visual (such as tables, figures, and graphs), will not be accessible to all users, even when supplemented with auditory descriptions. 

These findings align with established accessibility guidelines, such as the Web Content Accessibility Guidelines (WCAG) \cite{wcag} and the Accessible Rich Internet Applications (WAI-ARIA) standards \cite{world2023accessible}, which advocate for the use of multiple modalities to accommodate diverse user needs. Incorporating these principles and actively involving the disability community in designing and developing explainable AI approaches becomes essential for achieving meaningful accessibility and supporting informed, equitable user engagement in AI discussions.

\section{Conclusion}
\label{conclusion}
In conclusion, this study highlights the urgent need for accessible AI explanation systems that empower users, including those with disabilities such as sight loss, to understand and effectively interact with AI-driven decisions. This work was guided by two research questions: identifying gaps and challenges related to accessibility in current XAI approaches (RQ1), and exploring how AI explanations can be evaluated and understood by non-visual users (RQ2). Each component of the study was designed to address these questions and provide practical insights.

The literature review provided a comprehensive overview of the current landscape and directly informed RQ1 by revealing a strong reliance on visual explanation formats, which often exclude users with vision impairments. To explore RQ2, we proposed a four-part methodological proof of concept framework that included the categorization of AI systems, the definition of a user persona, the implementation of an accessible prototype, and the evaluation of this prototype with accessibility experts and users with lived experience of sight loss.

The evaluations revealed important insights. Simplified explanations seemed to be more effective for non-visual users than detailed ones, suggesting the value of offering multiple explanation formats tailored to diverse user needs. However, this finding should be interpreted with caution, as it is based on feedback from a small sample of only three participants. While these preliminary results reinforce the idea that a one-size-fits-all model of explanation is inadequate for inclusive AI systems, further research with a more diverse group of users is necessary to validate and generalize these findings.

Although this research is preliminary and limited by the specificity of the case study and the number of participants, it establishes a foundation for future work. Further research should explore additional multimodal explanation formats and engage broader user populations to develop more inclusive and generalizable solutions. This study contributes to the field by offering a structured approach for aligning AI decision-making with user understanding, emphasizing that transparency, accountability, and inclusivity are fundamental characteristics of responsible AI systems.

\bibliographystyle{named}
\bibliography{ijcai25}


\section*{Supplementary material}

# Supplementary Material

Table S1: Overview of recent studies evaluating XAI techniques. (+) indicates a new XAI technique; (*) denotes accessibility consideration.

| Ref | XAI techniques | Evaluation method | Metrics | Evaluators |
| --- | --- | --- | --- | --- |
| Aliyeva and Mehdiyev [2024] | SHAP, ICE, CFE | Multi-criteria decision analysis (MCDA) with Z-numbers-based Weighted Sum Model (WSM); Interviews | AUROC, Accuracy, Sensitivity, Specificity, F-measure, Matthews Correlation Coefficient (MCC) | 2 senior doctors, 1 general physician, and 1 resident doctor |
| Famiglini *et al.* [2024] | Class Activation Maps (CAMs) | Between-subject and within-subject study designs | Diagnosis accuracy, Confidence levels, Subjective feedback on usefulness | 8 spine surgeons and 8 musculoskeletal radiologists |
| Jin *et al.* [2024] | SmoothGrad | Clinical study evaluating AI-assisted glioma grading | Accuracy and p-value | 12 attending neurosurgeons, 2 neurosurgical fellows, and 21 neurosurgical residents |
| Sivaprasad *et al.* [2024] | Decision Trees, SHAP | Task-based evaluation with end-users | Completeness, Understandability, Explanation verboseness, Change in the mental model and Error in understanding | 50 non-expert end-users |
| Belardinelli *et al.* [2024] | AR-XAI, Gaze-Speech | User study with interaction interface | Behavioral measures, Subjective evaluation | 20 scientific and administrative staff |
| Jahn *et al.* [2024]+ | ACE, CARE | Functionally-grounded evaluation, Human-grounded evaluation, Case study | Sparsity, Density constraint, Abnormality, Perceived trust in AI system and Perceived helpfulness of explanations | 40 participants with a minimum level of primary school education |
| Makridis *et al.* [2024] | LIME, SHAP, Saliency Maps, Attention Mechanisms, Direct Trajectory Visualization and PFI | User survey | User preferences, Trustworthiness | Technical users and non-technical users (quantity not provided) |
| Gallée *et al.* [2024]+ | Proto-Caps | User study questionnaire | User performance, Trust in the model, Helpfulness of explanations, Misleading effects | 4 senior radiologists and 2 assistant physicians |

| Ref | XAI techniques | Evaluation method | Metrics | Evaluators |
|---|---|---|---|---|
| Labarta *et al.* [2024]* | LRP, Grad-CAM, LIME, SHAP, Integrated Gradients and Confidence Scores | Objective Human-Centered Evaluation (survey questionnaire) | Accuracy, Sensitivity, Specificity, Hypothesis Testing, Effect Size (Cohen's d) | 139 general users |
| Karagoz *et al.* [2024] | LIME, SHAP | Group user study | Trust, Confidence, accuracy | 97 medical experts |
| Rong *et al.* [2024]+ | I-CEE, Bayesian Teaching (BT) | Simulated experiments on four datasets, Human subject study | Simulatability accuracy, Human subjective understanding scores, Hypercorrection effect | 100 general users |
| Jansen *et al.* [2024]+ | Counterfactual, SHAP, Decision Trees, Visual Map | Human-grounded evaluation, Online experiment | Explanation Satisfaction Score, Cognitive Load, Overall Evaluation Score | 49 participants with varying levels of AI-literacy |
| Guo *et al.* [2024]+ | Counterfactual explanations | Algorithmic segmentation validation, Expert evaluation | Percentage of counterfactuals generated, plausibility, interpretability, and domain relevance | 1 pediatric cardiologist and 1 data scientist specializing in cardiac imaging |
| Lombardi *et al.* [2024] | SHAP, Counterfactual Explanations | Survey-based study | Correctness of user responses, Self-assessment scores, Trust and satisfaction scores | 16 neurologists |
| Morrison *et al.* [2024] | Natural Language Explanations, Example-Based Explanations | User study with interaction interface | AI reliance, accuracy | 83 experts and 53 non-experts |
| Schmitt *et al.* [2024] | Salient Feature, Free-Text Explanations | Crowdsourced user study | Performance, Understandability, Usefulness, Trust | 406 crowdworkers and 27 journalists |
| Yang *et al.* [2024] | Grad-CAM, LRP, Attention Rollout, Transformer Attribution | DeepFake Detection Task, Direct Evaluation, Indirect Evaluation, Human Annotations | Similarity Index, Pearson's Correlation Coefficient, User Voting Analysis, Confidence Test | 161 general users |
| Metta *et al.* [2024] | LIME, LORE, Saliency Maps, ABELE | Evaluation survey | Multi-class accuracy, Specificity, Recall, F1-score, Precision, Explanation effectiveness | 156 experts and non-experts |
| Del Águila Escobar *et al.* [2024]+ | Natural Language Explanations | User study via questionnaire to assess explanation comprehensibility and legibility | Percentage of correct responses in Cloze Test, Preference percentage in Binary Forced Choice test | 38 general users |

| Ref | XAI techniques | Evaluation method | Metrics | Evaluators |
|---|---|---|---|---|
| Cesarini *et al.* [2024] | LIME, SHAP, SAGE, BRCG, Anchors, Quantitative Input Influence, Decision Trees, Logistic Regression, Naive Bayes | Online user study | Usefulness, Satisfaction, Trustworthiness | 42 experts in XAI and non-experts |
| Nagy and Molontay [2024]* | LIME, SHAP, Permutation Importance, Partial Dependence Plot | User study to evaluate the interpretability, usefulness, and aesthetics of the XAI methods applied in dropout prediction | Interpretability score (Likert scale), Usefulness rating, Aesthetic appeal, Correct answer percentage in user study | 10 higher education decision-makers and 42 students from college and high school levels |
| Li *et al.* [2024] | SHAP | Human survey | User preference, comparison of trust in explanations, correlation analysis | 46 general users |
| Sovrano and Vitali [2023] | TreeSHAP, CEM | Multiple-choice quiz | Degree of Explainability (DoX) score | 192 general users |
| Naiseh *et al.* [2023] | Local explanations, Example-based, Counterfactual and Global explanations | Empirical study with 41 medical practitioners using clinical decision support systems | Trust calibration effectiveness | 41 Medical practitioners |
| Kim *et al.* [2023] | SHAP | Using the prototype and responding to an online survey, questionnaire of System Causability Scale (SCS) | Expert feedback on contextual XAI design usability, explanation quality | 5 brain-computer interface experts |
| Jmoona *et al.* [2023] | LIME, SHAP, DALEX | Quantitative and user evaluation exercise | Acceptability, Usability | 9 air traffic controllers |
| Röhrl *et al.* [2023] | LIME | User tests | Behavioral understanding, trustworthiness, bias detection | 57 biomedical and data scientists |
| Hu *et al.* [2023] | LIME, SHAP | User evaluation | User satisfaction, UI assessment | 103 AI non-experts and 120 unfamiliar with healthcare |
| Vilone and Longo [2023] | Argumentation graph and Decision Trees | Psychometric assessment through human feedback | Reliability, Validity, Comprehensibility | 89 users with experience in AI technologies |
| Falatouri *et al.* [2023] | SHAP | Workshop with practitioners | Willingness to use, trust in model | Heating appliances practitioners (quantity not provided) |
| Kenny *et al.* [2023]+ | CBR, Feature Highlighting, FAM, CAM, LIME, Superpixel-Based Explanations | User study with interaction interface | Explanation Fidelity, User Perception of Correctness, Testing Accuracy Drop, Human Judgment Calibration | 163 general users |

| Ref | XAI techniques | Evaluation method | Metrics | Evaluators |
|---|---|---|---|---|
| Roy *et al.* [2023]+ | LIME, SHAP, PKiL | Domain experts reviewing model explanations | Model Agreement, Accuracy, AUC-ROC Scores, Faithfulness of Explanations | Clinicians (quantity not provided) |
| Das *et al.* [2023]+ | LIME, SHAP, LIME+, Anchors | Expert analysis and user studies | Explanation sensibility, User preference, User confidence, Computational efficiency | 3 Machine learning experts and 161 non-expert users |
| Hernandez-Bocanegra and Ziegler [2023]+ | Interactive, conversational explanations, Intent-based explanatory, Argumentation theory-based explanations | User study with interaction interface | Helpfulness, Comparison of explanation effectiveness, perceived explanation quality | 223 general users |
| Cau *et al.* [2023] | Logic-style explanations | Human-grounded evaluation | Trust, Confidence, Preference and Perceived accuracy | 1324 general users |
| Schmude *et al.* [2023] | Textual, Dialogue, Interactive | Qualitative task-based study | Fairness perceptions | 30 general users |
| Meza Martínez *et al.* [2023] | Anchors, DICE, LIME, SHAP | Evaluated explanations using eye-tracking technology, self-reports, and interviews | Trust, understandability, usefulness, satisfaction | 2 data scientists, 3 HCI researchers and 506 general users |
| Morrison *et al.* [2023] | LIME, Grad-CAM, Feature Visualization | Game With a Purpose (GWAP) called Eye into AI | Agreement, Exposure, Copiousness | 50 general users |
| Lim and Perrault [2023] | LIME, SHAP, Attention Mechanism, Comments-based, Evidence-based | Experimental user study | Agreement with the AI's veracity predictions, Intent to share news, Variance in agreement scores, usability, trust, and perceived effectiveness | 180 general users |
| Colley *et al.* [2023] | Feature Relevance, Local Explanations | Use cases, such as recipe suggestions and jogging route recommendations | Participant understanding, Perceived trust, Effectiveness | 10 general users |
| Meza Martínez and Mädche [2023] | SHAP | Semi-structured interviews | User feedback, usefulness, user understanding | 13 bachelor's, 7 master's, and 1 PhD student |
| Boidot *et al.* [2023] | Confidence Score, Surrogate Rule-based, LIME, Nearest Neighbor-based | User study - Participants made decisions with different XAI explanations | Decision Accuracy, Decision Time, Use Rate, Perceived Usefulness, Interpretability Score | 27 students and young engineers |
| Cabitza *et al.* [2023] | Text-based explanations | A questionnaire-based user study | Trust, comprehensibility, appropriateness, and utility, Diagnostic accuracy | 44 cardiologists |

| Ref | XAI techniques | Evaluation method | Metrics | Evaluators |
|---|---|---|---|---|
| Ibrahim *et al.* [2023] | SHAP, DiCE | Online experiment | Prediction accuracy, Performance gain, Influence score, Bias measures, Self-reported trust and confidence | 559 general users |
| Malandri *et al.* [2023]+ | LIME, SHAP | Online user experiment | Explanation completeness, precision, granularity, causality, ease of understanding, clarification | 15 technical participants, 15 non-technical participants and 15 managers |
| Metta *et al.* [2023] | IntGrad, GradInput, LRP, LIME, SHAP, ABELE | User survey | Classification Accuracy, Confidence Ratings, Likert-scale ratings to assess the usefulness | 156 participants (94% with a scientific background, including 27% in medicine or dermatology) |
| Förster *et al.* [2023]+ | Counterfactual explanations | User study assessing explanations generated by the proposed method | Realism and typicality, feasibility, smoothness, sparsity, validity | 174 general users |
| Veldhuis *et al.* [2022]+* | Counterfactual Explanations, SHAP, ReCo | Quantitative evaluation and qualitative feedback | Realism score, proximity, sparsity, validity | 8 DNA experts |
| Mualla *et al.* [2022]+ | Contrastive Explanations, Explanation Filtering | Empirical user study with parametric and non-parametric statistical testing | Trustworthiness, Understandability, Satisfaction (Likert scale) | 90 general participants |
| Dieber and Kirrane [2022] | LIME | Usability study and performance assessment of LIME on tabular models | User experience and usability | 12 academics or pursuing a degree |
| Schulze-Weddige and Zylowski [2022] | SHAP, LIME, alibi explain | Interviews | Understandability, interaction, preference | 15 non-experts |
| Kim *et al.* [2022] | Grad-CAM, BagNet, ProtoPNet, ProtoTree | Self-rated understanding of the XAI methods before and after tasks | Mean task accuracy, standard deviation and p-value | 1000 general users |
| Warren *et al.* [2022] | Counterfactual and Causal explanations | User study with training and testing phases | Objective accuracy, Subjective satisfaction, trust, understanding of features | 127 general users |
| Bove *et al.* [2022] | SHAP | Experimental conditions | Objective understanding score, Satisfaction rating, ANOVA statistical analysis | 80 non-expert users |
| Xu [2022]+ | Dialogue-based explanation | User study with interaction interface | User preference, Ease of understanding, explanation clarity | 24 general users |
| Arrotta *et al.* [2022a]+ | Grad-CAM, LIME, Model Prototypes | User-based survey | Explanation Score, User satisfaction, accuracy and F1-score | 147 non-expert users |

| Ref | XAI techniques | Evaluation method | Metrics | Evaluators |
|---|---|---|---|---|
| Wang and Yin [2022] | Feature importance, Feature Contribution, Nearest Neighbors, Counterfactual Explanations | AI-assisted decision making tasks | Objective understanding, Subjective understanding, Reliance on the model, Trust calibration | 2,008 participants with varying levels of domain expertise in decision-making contexts |
| Aechtner *et al.* [2022] | LIME, SHAP, PDP | Users evaluated AI explanations related to their likelihood of acceptance, Survey-based methodology | Trustworthiness, usefulness, informativeness, satisfaction, and understandability | 60 University students (AI novices (60%) and AI experts (40%)) |
| Ma *et al.* [2022] | PFI, SHAP, DiCE | Web-based user study | User Accuracy, Understanding Score, Explanation Request Rate, User Satisfaction Score | Over 200 non-expert users |
| Arrotta *et al.* [2022b]+ | Grad-CAM, LIME, Model Prototypes | Comparison of three different XAI methods | Explanation Score, User satisfaction, Accuracy and F1-score | 121 users (26% high school, 58% bachelor's degree, 11% master's degree, 5% PhD) |
| Antoniadi *et al.* [2022] | SHAP | Short survey with users | Trustworthiness, explainability, and usefulness | 8 healthcare professionals |
| Meas *et al.* [2022] | SHAP | User feedback on the generated explanations and their usefulness in decision-making | Feature importance and contribution scores from SHAP explanations, user ratings and satisfaction levels | 7 expert engineers |
| Singh [2022] | Layerwise Relevance Propagation (LRP) | User questionnaire survey | Interpretability, Trust, Relevance score analysis, Saliency comparison | 19 university students |
| Van Der Waa *et al.* [2021] | Contrastive rule- and example-based explanations | Interaction with experiments and questionnaires | Understanding of the system, persuasiveness of the explanation and task performance | 90 participants from lower vocational to university education |
| Kenny *et al.* [2021]+ | COLE-HP (Contributions Oriented Local Explanations - Hadamard Product) | Three user experiments evaluating correct/incorrect classifications and different error rates | Prediction accuracy, reasonableness, confidence and user satisfaction | 514 general users |
| La Gatta *et al.* [2021b]+ | LIME, PASTLE | Comparative performance assessment with real-world datasets and user studies | Explanation effectiveness, Model interpretability, User trust | 36 users with basic knowledge about Machine Learning |

| Ref | XAI techniques | Evaluation method | Metrics | Evaluators |
|---|---|---|---|---|
| Moradi and Samwald [2021]+ | MUSE, LIME, CIE | Comparative analysis of XAI techniques on tabular and textual classification tasks | Accuracy and interpretability | 20 researchers or postgraduate students in AI, ML or related fields |
| La Gatta *et al.* [2021a]+ | CASTLE, Anchors | Empirical evaluation on datasets, user tests for qualitative evaluation | Interpretability | 32 undergraduate students familiar with ML or statistics |
| Chromik [2021]+ | SHAP, LIME, SHAPRap | Formative user study | Interpretability, user preference | 16 participants with at least a graduate degree |
| Schneider *et al.* [2021] | Grad-CAM | Two text classification tasks | Decision Fidelity, Explanation Fidelity, Detection Accuracy, User Agreement Score | 140 participants (16% high school, 11% associate degree, 56% bachelor's degree, 16% master's degree, 1% doctoral degree) |
| Wang and Yin [2021] | Feature Importance, Feature Contribution, Nearest Neighbors, Counterfactual Explanations | Recidivism prediction and forest cover prediction tasks | Understanding AI Models, Uncertainty Awareness, Trust Calibration | 1,343 general users |
| Mucha *et al.* [2021] | LIME | Online experiment | Weight of Advice, Estimation Errors, explanation usefulness | 60 participants from a secondary school |
| Jesus *et al.* [2021] | LIME, SHAP, TreeInterpreter | Real-world decision-making setting | Accuracy, Recall, False Positive Rate (FPR), Decision Time, Agreement (Fleiss' Kappa), explanation usefulness, relevance, and diversity | 3 fraud analysts |
| Leung *et al.* [2021] | SHAP, Contrastive Explanations, Decision Trees | Case study assessing the usefulness and ease of understanding of the explanations | Ease of understanding, Perceived usefulness, Prediction Accuracy, Explanation Accuracy | 15 general users |
| Gupta *et al.* [2021] | Local Explanations | Ranking images based on visual explanations | User Satisfaction, Usefulness of Explanations | 20 general users |
| Alipour *et al.* [2020]+ | Spatial Attention, Active Attention, Bounding Box, Scene Graph, Object Attention, Textual Explanations | Prediction evaluation task | Accuracy, Trust, Confidence, Chi-Square Test Analysis | 90 general users |

| Ref | XAI techniques | Evaluation method | Metrics | Evaluators |
|---|---|---|---|---|
| Zhou *et al.* [2020] | Interactive Alchemy Explainable, Static Alchemy Explainable | A pretest to assess prior probability knowledge, One of the two explainables (interactive or static), A posttest to measure learning gains, inspired by Bloom's Taxonomy | Learning Residual Score, learning outcomes, effectiveness of explanations | 88 participants with undergraduate backgrounds in computer science |
| Spinner *et al.* [2019] | LIME, ANCHORS, CAVs, LRP, Deep Taylor Decomposition, Saliency Maps, Gradient-based explanations, DeConvNet | Evaluation of a framework's effectiveness | Participant feedback, trust, effectiveness, Observations of user interactions and workflow patterns | 9 participants with varying expertise levels, from model novices to model developers |
| Gunning and Aha [2019] | Deep Explanation, Interpretable Models, Post-hoc Explanations (LIME, SHAP), Case-Based Reasoning, Bayesian Rule Lists | Human-in-the-loop psychologic experiments | User Satisfaction, Mental Model Understanding, Task Performance, Explanation Fidelity, Appropriate Trust and Reliance | Naval Research Laboratory |

Table S1: Overview of recent studies evaluating XAI techniques. The symbol (+) marks studies that proposed new XAI techniques, and (*) indicates that the study considered any accessibility issue.

## Table S2: Types of AI: Categories, Definitions, and Examples

| Category | Definition | Example |
|---|---|---|
| **Impact-Centric AI** | | |
| User-Centric | Refers to AI that makes decisions or suggestions directly impacting the user interacting with the system. Vereschak *et al.* [2024] | When using an AI-powered fitness app, the system tracks the user's physical activity, progress, and preferences to deliver personalized workout recommendations and dietary suggestions tailored to their goals. |
| Third-Party | Refers to AI that makes decisions or suggestions affecting third parties who do not directly interact with the system but are impacted by the actions of another user. Vereschak *et al.* [2024] | PredPol is a predictive policing program that utilizes historical crime data to distribute police resources. In this context, a police officer can use the system to determine where to deploy cops and how many to assign. O'Neil [2016] |
| **Function-Based AI** | | |
| Predictive | Uses historical data to predict future events or trends (what is most likely to happen in future?). Bokonda *et al.* [2020] | Sales forecasting, financial decisions and risk analysis. |
| Generative | In simple terms, "generative" implies "generate", "produce", or "create" new data or content, such as images, text, music, or code, based on patterns learned from training data. Kalota [2024] | Digital art generation, text creation and voice synthesis. |

| Category | Definition | Example |
|---|---|---|
| Descriptive | Analyzes historical data to identify patterns and trends, helping to understand what happened (what has happened in past?). Roy *et al.* [2022] | Sales data analysis, performance reports, and analytical dashboards. |
| Prescriptive | Recommends actions based on predictive and descriptive data, suggesting the best courses of action to achieve specific goals (What can be done to achieve a certain goal?). Roy *et al.* [2022] | Medical treatment recommendation systems, logistics and supply chain optimization, and personalized marketing. |
| Task-Based | AI systems that are developed to perform specific tasks, either in a single domain or across multiple domains. Gutierrez *et al.* [2023] | GPT-3 was initially trained to predict the next word in a text string. It has since been fine-tuned to support new tasks like translation and coding. Kalyan [2023] |
| Pattern Recognition | Identifies patterns or regularities in large datasets. Kim [2010] | Facial recognition, image classification, and speech recognition. |
| Interactive | Interacts with users adaptively and responsively, often used in user interfaces and virtual assistants. Raees *et al.* [2024] | Chatbots, virtual personal assistants, and dialogue systems. |
| **Transparency-Based AI** | | |
| White-Box | Refers to AI models with decisions and processes that can be understood and explained, where the internal logic and decision criteria are accessible and auditable. Wanner *et al.* [2020] | Decision trees, linear regression, and rule-based systems. |
| Gray-Box | Models that offer some transparency but still have opaque or complex parts that are difficult to interpret. Wanner *et al.* [2020] | Neural networks with partial explainability, and hybrid models. |
| Black-Box | Models with decisions and internal processes that are opaque and difficult to interpret. Wanner *et al.* [2020] | Deep neural networks, XGBoost, and deep reinforcement learning systems. |
| **Reasoning-Based AI** | | |
| Deductive | Deductive reasoning is a type of logical thinking that starts from general principles and moves towards specific instances, ensuring that the conclusion is true if the premises are true. Williamson *et al.* [2002] | Rule-based systems such as logical inference engines and rule engines. |
| Inductive | Refers to an AI's ability to make generalizations from specific data. This process involves creating hypotheses or inferring general rules based on concrete examples. Chater *et al.* [2011] | Machine learning algorithms, such as neural networks and decision trees. |
| Abductive | Proposes the best possible explanation for a set of observations. It is often used for diagnosing and understanding problems where causes are not clearly known. Liang *et al.* [2022] | Medical diagnostic systems that infer the most likely disease from presented symptoms. |
| Analogical | Analogical reasoning involves drawing parallels between two domains or situations by identifying a similar relationship or structure. It relies on transferring knowledge from a familiar situation (the source) to a less familiar one (the target). Sowa and Majumdar [2003] | Recommendation systems that suggest items based on similar user preferences. |

| Category | Definition | Example |
|---|---|---|
| Statistical | Uses statistical methods to infer the probability of certain events or outcomes based on historical data. It is widely used in AI for modelling uncertainty and decision-making under uncertainty. Garfield [2002] | Probabilistic models such as Bayesian networks and stochastic processes. Political analysis, which studies and interprets polls and elections. |
| Case-Based | Solves new problems by recalling and adapting solutions from previously stored similar cases. Jian *et al.* [2015] | Decision support systems that use databases of past cases to suggest solutions. |
| Non-Monotonic | Allows conclusions to be withdrawn or revised as new information is received. It is helpful in dynamic domains where rules and knowledge may change. Przymusinski [1988] | A recommendation system suggests movies based on a user's preferences. If the system learns that the user no longer enjoys horror movies, it will stop recommending horror movies in the future, even if the user used to like that genre. |

Table S2: Types of AI: Categories, Definitions, and Examples


\end{document}